# Offline Handwritten Chinese Text Recognition with Convolutional Neural Networks

Brian Liu, Xianchao Xu, Yu Zhang
Intel Corp.
Beijing, China
{bo.a.liu, xu.james, richard.yu.zhang}@intel.com

*Abstract*— Deep learning based methods have been dominating the text recognition tasks in different and multilingual scenarios. The offline handwritten Chinese text recognition (HCTR) is one of the most challenging tasks because it involves thousands of characters, variant writing styles and complex data collection process. Recently, the recurrent-free architectures for text recognition appears to be competitive as its highly parallelism and comparable results. In this paper, we build the models using only the convolutional neural networks and use CTC as the loss function. To reduce the overfitting, we apply dropout after each max-pooling layer and with extreme high rate on the last one before the linear layer. The CASIA-HWDB database is selected to tune and evaluate the proposed models. With the existing text samples as templates, we randomly choose isolated character samples to synthesis more text samples for training. We finally achieve 6.81% character error rate (CER) on the ICDAR 2013 competition set, which is the best published result without language model correction.

## I. INTRODUCTION

Text recognition is one of the most essential parts for artificial intelligence. With the rise and popularization of deep learning technology, multilingual text recognition which includes printed text recognition [2], handwritten text recognition [3] and scene text recognition [4] are being actively researched and have achieved breakthrough results.

Most of the state-of-the-art works [2, 4, 5, 6] use hybrid CNN and RNN or LSTM architectures to solve these text recognition tasks. In order to reduce the model size and speed up the training, and also reserve the performance, there is a recent shift to recurrent-free architectures [7, 8, 9] for English-centric text recognition tasks due to the better parallelism of convolutional networks. Both of the [7, 8] actually introduced the similar gating mechanism of LSTMs to connect layers in convolutional networks. While in the [9], only convolutional layers are used without recurrence for scene text recognition, which implies that the convolution could still model interaction between characters in text.

Given the thousands of vocabularies and great diversity of writing style, the handwritten Chinese text recognition (HCTR) is still a big challenge. Over the last decade, there are many different and efficient approaches for HCTR, thanks to the well-known CASIA-HWDB [1] database. [10] presents an over-segmentation approach which treats the HCTR as two-stage task. It consists of steps of over-segmentation of a text line image, construction of segmentation and character candidate lattices, and path search from these lattices with context information. Although the hybrid language model play important roles for the result, the segmentation before recognition will brings in extra errors. [11] proposes a writer-aware CNN based on parsimonious HMM to tackle the vocabulary and writing style issues for HCTR. Its adaptive training approach and compact design of HMM plus with hybrid language model contribute to state-of-the-art result. But the multi-pass decoding strategy in testing stage will cost too much time. While in the [12, 13], the authors show that end-to-end method with CNN and multi-layered residual LSTM and CTC or ACE [13] exhibit promising performance and even superior performance over attention related mechanism. They also mention that their model have achieved state-of-the-art results on the popular ICDAR 2013 competition set of CASIA-HWDB without or with language model.

In this work, we use a different approach for the HCTR challenge. Similar to the [9, 14], we choose the simple and end-to-end CNN + CTC method for sequence modeling. This method is first used not only in the scene text recognition in the [9], but also in the speech recognition in the [14]. It is proved to be working well, with comparable performance as other LSTM-based systems. Back to HCTR, there is no other published work using this CNN + CTC method.

We start from the latest version of PyTorch [15] and the build-in convolutional networks. As most of the previous works did, we treat the text recognition as image level sequence modeling task. The CTC loss function we adopt is Baidu implemented warp-ctc[1]. Considering the challenging of HCTR, we use well-known VGG-16 [16] as the baseline. To further improve the performance, we ease the overfitting problem by introducing the dropout [17] strategy. We place the dropout layer after each of the max-pooling layer and apply very high dropout rate especially on the input of linear layer, which is proved to be very effective. We also upgrade our model to residual connection based networks [18] and provide a set of network configurations with different blocks for model size and performance trade-off. In our experiments, we find that the characters in the text samples could be appeared in random order instead of the order in normal corpus. With this assumption, we construct more training samples with isolated character database of CASIA-HWDB and we achieve even higher performance. Based on our results, we argue that the recurrent connection might not be the first choice for text recognition, since the input vectors are in pixel-column format but not word vectors as NLP task, especially for the complex characters which span tens of columns. The mainly contributions are summarized as followings:

---
[1]A PyTorch binding at https://github.com/baidu-research/warp-ctc.

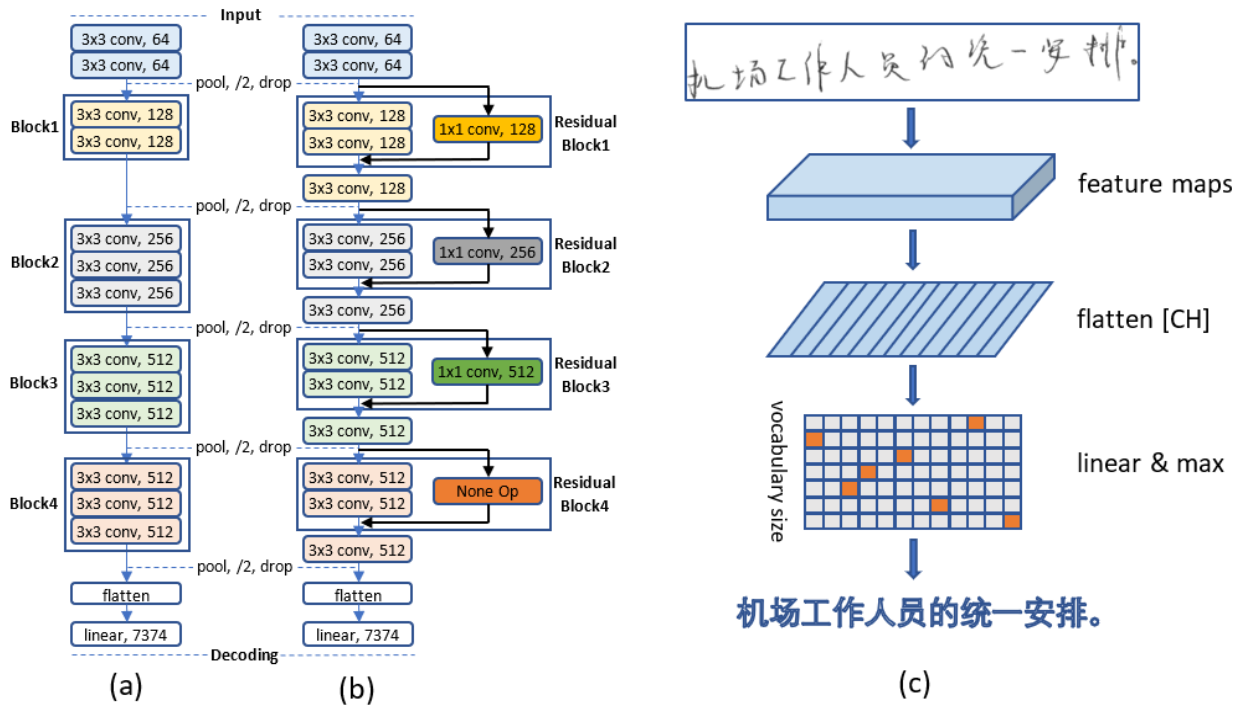

Figure 1: Overview of the convolutional neural network model architecture for HCTR. (a): Standard VGG-16 network without fully-connection layers for feature extraction, and it is followed by a flatten (in C&H dimension) layer for being connected to final linear layer. Max-pooling and dropout layers are added after each block. (b): Upgraded network with residual blocks and one additional convolution layer after each block. (c): Network output is simply decoded by translating the index with maximum value at each sequence step and removing the "[blank]" and repeated character as defined by CTC.

- The CNN+CTC method is first used for HCTR and it can be easily implemented and trained efficiently.
- Dropout is introduced and applied after each of the max-pooling layer, and very high dropout rate on the input of linear layer improves the performance in a great level.
- Synthesizing new text with random order characters opens the way to increase the amount of data for text recognition model training.

To demonstrate the performance, we evaluate our best model on the popular ICDAR 2013 competition set of CASIA-HWDB and achieve the best result in the benchmark of CER without language model.

The rest of the paper is organized as: Session II gives the overview of the model architecture and related techniques. In Session III, we will describe the optimization of the model on target database including the data construction details. And the experiments and results are shown in Session IV. The last session concludes the paper.

## II. MODEL ARCHITECTURE

We simply use only convolutional layers to extract features from images, and the CTC (connectionist temporal classification) loss function to learn the classification of those features into each individual character. Fig 1. illustrates the overview of the proposed model architectures.

### A. Convolutional Neural Networks

The simple and relative shallow VGG-based models have achieved promising performance for broad range of computer vision tasks. And in most of recent models for text recognition also include the similar structure to stack layers for feature extraction. The model architecture in Fig. 1 (a), reuses all the convolution layers from the official VGG-16 [16] network except the fully-connection layers. Batch normalization [19] and relu layer follow each of the convolution layer, and max-pooling and dropout layer follow each block with same output channels. After these convolutional layers, the 3D (C*H*W) dimensional features are flatten to 2D in [C and H] (or [H and W] as optional) dimension for being connected to the last linear layer for prediction per class.

Residual connections can ease the training of networks that are substantially deeper than those used previously [18]. Official ResNet architecture is mainly designed for ImageNet challenges with coarse-grained objects in 224*224 resolution images, while for the fine-grained characters in images with lower height and much longer width it does not work that well. In the Fig. 1 (b), the proposed architecture is the combination of partial VGG-16 network and residual connections. All the internal structure of these four residual blocks reserves the same as official ResNet implementation but with different number of output feature maps and the stride size which is now 1 instead of original 2. And an additional convolution layer is added after each of the residual block inspired by another Resnet-based architecture [20] which is relatively effective for scene text feature extraction.

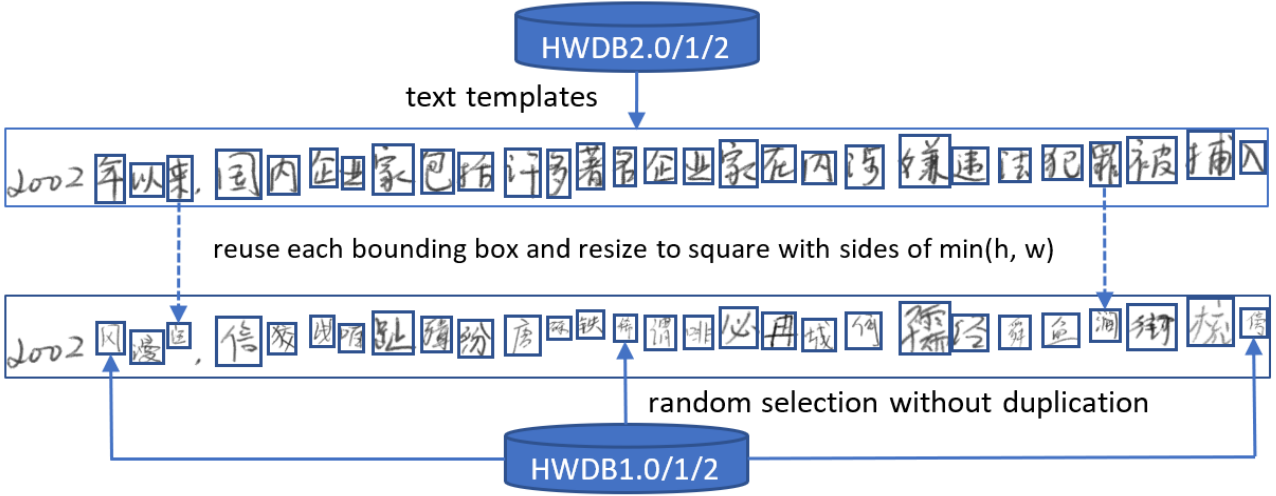

Figure 2: New text samples synthesis process from isolated characters database.

## B. Connectionist Temporal Classification

The paper [21] presents this CTC technique for training RNNs to label un-segmented sequences at first, and then from speech recognition to text recognition tasks and from RNNs-based networks to CNNs-based networks, this method is well received and adopted. It eliminates the need to segment the sequences and opens the way to collect more data for date-driving training. The basic idea behind CTC is to interpret the network outputs as a probabilities distribution over all possible label sequences (alignments), conditioned on a given input sequence. Additional blank label is introduced to conduct valid alignments and the output parse. Given long sequences in text and big vocabulary set, the amount of alignments would be huge and so be the computation. A dynamic programing algorithm solves this challenge well which is detailed in the original paper. During the inference stage, decoding the network outputs to readable transcription can be taken by greedy search or beam search combined with linguistic context, etc. In the paper, we simply apply the greedy decoding method for computation efficiency.

## III. MODEL OPTIMIZATION

In order to improve the model performance, some common and effective methods are leveraged and evaluated empirically. Differentially, we propose fixed-height and fixed-ratio image resizing with normalization right-side padding strategy. We apply dropout on the input of both some convolution layers and liner layer with optimized dropout rates. And for data synthesis, we directly reuse the existing text samples as templates and fill with characters in random order.

### A. Input Preprocessing

With the constraints of the some layers in the leading deep learning frameworks, most of the text recognition models use the fixed-height image as input. In real world, the captured images can be in any resolution. Resizing images to fixed-length [22, 4, 12] is popular for implementation simplicity, but images with too short or long width will get much distorted in character level. We in turn adopt the fixed-height and fixed-ratio resize method at the same time on gray-scale images, so after the resizing the input image will have dynamic length, similar as the way in [2]. During the padding phase for batch mode training and testing, we move a further step to pad each target region using the last right column of input which is normalized in advance [4].

### B. Dropout after Max-Pooling

The dropout [17] which is a simple way to prevent neural networks from overfitting has been integrating into many networks for supervised learning tasks. At the beginning of these deployments, the dropout layers are mainly inserted among fully-connection layers, for example, in AlexNet [23], VGG [16], etc. And then the dense blocks in DenseNet [24] also add the dropout after each convolution layer but before the batch normalization layer. More recent researches [6, 26] show that placing the dropout directly on the convolution layers seems better. In this work, we just place the dropout after each max-pooling layer for simplicity. In the Tab. 1, we provide an exploration about the dropout rates for our VGG-16 baseline model. We found that the dropout strategy in deed helps improve the performance in a great level (-3.24%). And the biggest improvement (from 13.61% to 11.04%) comes from the very high rate (0.9) on linear layer, this might because the size of linear output which is also the number of vocabulary set is huge for this HCTR special case. We also found that the drop rate applied on the input of first block has relatively little impact on the result. In our later experiments, we choose the 12$^{th}$ row in Table 1 as the optimized drop rates configuration.

Table 1: Dropout rates exploration on the baseline model.

| ID | Dropout rates on input of Blocks and Linear | | | | | CER (%) |
|---|---|---|---|---|---|---|
| | Block 1 | Block 2 | Block 3 | Block 4 | Linear | |
| 1 | 0.0 | 0.0 | 0.0 | 0.0 | 0.0 | 13.61 |
| 2 | 0.0 | 0.0 | 0.0 | 0.0 | 0.5 | 12.42 |
| 3 | 0.0 | 0.0 | 0.0 | 0.0 | 0.6 | 11.66 |
| 4 | 0.0 | 0.0 | 0.0 | 0.0 | 0.7 | 11.61 |
| 5 | 0.0 | 0.0 | 0.0 | 0.0 | 0.8 | 11.20 |
| 6 | 0.0 | 0.0 | 0.0 | 0.0 | 0.9 | 11.04 |
| 7 | 0.0 | 0.0 | 0.0 | 0.5 | 0.9 | 10.97 |
| 8 | 0.0 | 0.0 | 0.5 | 0.5 | 0.9 | 10.78 |
| 9 | 0.0 | 0.5 | 0.5 | 0.5 | 0.9 | 10.57 |
| 10 | 0.5 | 0.5 | 0.5 | 0.5 | 0.9 | 10.64 |
| 11 | 0.0 | 0.4 | 0.4 | 0.4 | 0.9 | 10.41 |
| 12 | 0.0 | 0.3 | 0.3 | 0.3 | 0.9 | **10.37** |
| 13 | 0.3 | 0.3 | 0.3 | 0.3 | 0.9 | 10.45 |
| 14 | 0.0 | 0.2 | 0.2 | 0.2 | 0.9 | 10.48 |

*C. Data Synthesis*

Despite of segmentation-free method in this paper, and no need to provide character level segmentation for training, given so much vocabularies and writing styles, the amount of existing data is still limited. There is also an extra requirement on the new data which should be follow the similar data collection and distribution as original data. Thanks to the well-known CASIA-HWDB [1] databases, both the text version HWDB2.x and isolated characters version HWDB1.x are compatible with each other. An intuitive way for data synthesis is to construct new text data from characters samples. In this paper, we propose to reuse the pre-labelled bounding box of each character in text samples and reserve the nature writing habits of people, and just replace the characters in HWDB2.x text with isolated characters from HWDB1.x. The most important difference of our data synthesis from others [26, 27] is that we select the characters in random order and do not use on any corpus. Fig.2 shows the detail process of data synthesis.

## IV. EXPERIMENTS

*A. Datasets*

We use the research only Chinese handwritten text database shared by CASIA-HWDB [1] to conduct out experiments and target its ICDAR2013 competition. The basic training set for this work is the offline handwritten Chinese text databases (HWDB2.0, HWDB2.1, and HWDB2.2), which are wrote by 1,020 people and include 52,230 text samples. With above data synthesis method, we generate more 173,383 text samples from the isolated offline handwritten Chinese character databases (HWDB1.0, HWDB1.1, and HWDB1.2), which are also wrote by those people. The size of vocabulary set our models support is 7373 which represents the number of all character class in the training set. And the evaluation set is the same as the ICDAR2013 competition set with 60 new writers and 3,432 text samples. Considering that the punctuations, numbers and English letters are originally labelled using Chinese input method in the ground truth of evaluation set, but not exist in the vocabulary set, we thus correct these characters in ground truth label files with the same ones but using English input method. In our all experiments, we use the character error rate (CER) and it is calculated using standard edit distance library. In the Eq.1, *Ni*,

$$CER = \frac{N_i + N_d + N_s}{N} \quad (1)$$

*Nd* and *Ns* are the number of insertion error, the deletion error and the substitution errors, respectively.

*B. Implementation details*

Our proposed models work with latest version of PyTorch [15] framework and are trained on single Nvidia TITAN V GPU. We choose 128 pixels as fixed height of input images for model training considering the complexities of Chinese characters, and this number can also be scaled down for performance and accuracy trade-off. The optimizer, momentum and weight decay are set to SGD, 0.9, and 1e-4 respectively by default. We mainly tune the batch size and learning rate hyperparameters. Since the average length of text samples is about in 1500 pixels wide in the training set, big batch size will consume almost all the graphic memory. Empirically, for our target GPU we set the batch size to 4 and initial learning rate to 1e-5 accordingly because of the limitation of batch normalization [19]. Learning rate will be adjusted by timing 0.1 if test accuracy does not increase within 10 epochs on basic training data and 5 epochs on basic plus synthetic training data. Generally, these models training will converge after about 50 epochs and 25 epochs respectively.

Table 2: ResNet-based networks architecture exploration and training with basic and synthetic data.

| Residual block configurations | Training data | Dropout rates on input of Residual Block and Linear | CER (%) |
|---|---|---|---|
| ResNet-1111 | Basic HWDB2x | 0.0-0.3-0.3-0.3-0.9 | 10.39 |
| ResNet-1221 | Basic HWDB2x | 0.0-0.3-0.3-0.3-0.9 | 9.79 |
| ResNet-1331 | Basic HWDB2x | 0.0-0.3-0.3-0.3-0.9 | 9.58 |
| ResNet-2451 | Basic HWDB2x | 0.0-0.3-0.3-0.3-0.9 | 9.41 |
| ResNet-2451 | Basic HWDB2x + Synthetic HWDB1x | 0.0-0.3-0.3-0.3-0.9 | **6.81** |

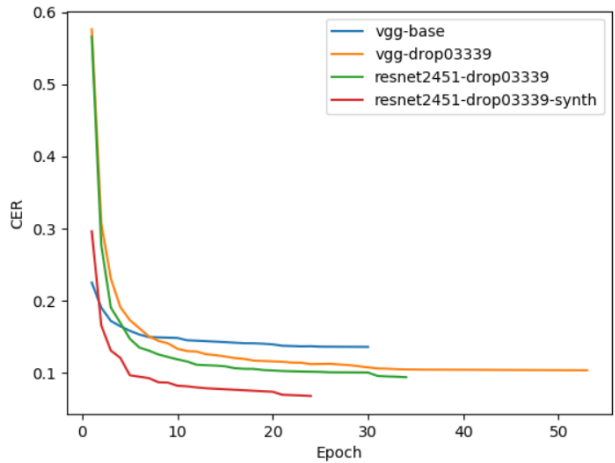

Figure 3: Training (epochs) and testing (CER) curve.

For quicker training on target device, we integrate the Nvidia apex [28] library which enables the mixed precise data format training and reserves the accuracy between original FP32 and FP16. As the simplicity of our models, we set the opt level as 2 and achieve 2-3x speed up without accuracy drop.

*C. Results and Analysis*

After explored the best configuration of VGG baseline model in Tab.1, we further focus on the upgraded residual networks architecture exploration with the same dropout strategy for even better performance. Because in our proposed models in Fig. 1 (b) the last two residual blocks have the same number of filters, we only adjust the number of first three blocks. Tab.2 lists the results of networks architecture exploration with different residual block configurations and training data. As expected, the CER decreases along with the addition of block number. Different block configurations aim to the flexibility of performance, model size and accuracy trade-off during the deployment phase on specific hardware devices.

Compared to the baseline model, the best residual model ResNet-2451 helps reduce the CER from 10.37% to 9.41%, and after trained with both basic and synthetic data, it further reduces the CER for another 2.6%. The last row in Tab.2, the 6.81% CER is our best result and also the best published result on the ICDAR 2013 competition set of CASIA-HWDB, to be the best of our knowledge. Combined with all above models, we draw out the training and testing curve in Fig.3 for visualizing the improvement of each optimization step.

Table 3: Comparison with recent HCTR works on benchmark of CER (100% - AR).

| Method | Vocabulary | Without LM | With LM |
|---|---|---|---|
| SMDLSTM-RNN + CTC [26] | 7356 | 13.36 | 9.62 |
| | 2672 | 9.98 | 7.39 |
| Seg. + Cascading CNN [29] | 7356 | 11.21 | 5.98 |
| Seg. + CNN + Hybrid LM [10] | 7356 | - | 3.68 |
| CNN-ResLSTM + ACE [13] | 7357 | 8.75 | 3.30 |
| WCNN-PHMM [11] | 3980 | 8.64 | 3.27 |
| | 7360 | 8.42 | 3.17 |
| CNN + CTC (This work) | 7373 | **6.81** | - |

In Tab.3 we make a comparison between our proposed method and recent HCTR research works on the same ICDAR2013 competition set with and without language model. Obviously, our CNN + CTC method outperforms the previous over-segmentation methods, LSTM-RNN based methods, and HMM-based methods with a great margin. Meanwhile, our model is trained end-to-end and built with recurrent-free architecture, thus our model is also most suitable for deployment with high performance. On the other hand, our result demonstrates that CNN itself is very powerful to represent a model with huge number of classes and CTC performs well to classify each of these classes. Note that, as these methods all integrate a language model and get about 5% performance boost, we can also add it and even better result could be expected.

Finally, in Fig.4 and Fig.5, some correct recognition samples and misrecognized samples are listed respectively. After analyzing the results, we found that there are major two kinds of difficulties leading to high errors: the extremely scribbled handwriting and distorted lines of text. We also provide the prediction results of all the 3432 samples in ICDAR2013 competition set using our best model for readers, at https://github.com/brianliu3650/OCRtextline-HCTR.

## V. CONCLUSION

In this paper, we propose a simple and effective method with CNN-only network for the challenging offline handwritten Chinese text recognition task. We iteratively improve the performance by customizing the baseline VGG model with residual connections, applying dropout on input of some convolution layers after each max-pooling layer and with very high rate on linear layer for huge Chinese vocabulary size. Data synthesis with random isolated characters and existing templates is easy-to-operate and plays an important role in data augmentation for text model training. Our mothed demonstrates its great superiority over other state-of-the-art approaches. Nevertheless, we are open to the more complex convolutional neural networks architecture and optimal dropout strategy. We also notice that there are much space to search the hyperparameters used for training, as the batch size is only set to 4 in this work. Applying group normalization [30] should be a good next step.

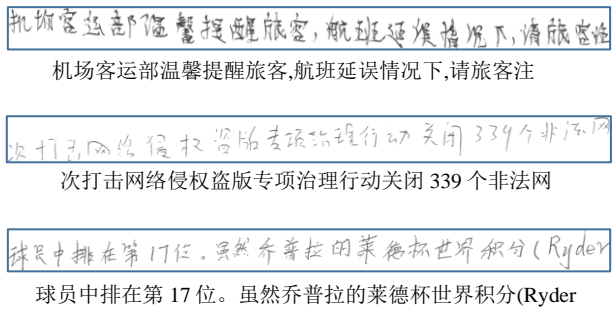

机场客运部温馨提醒旅客,航班延误情况下,请旅客注

次打击网络侵权盗版专项治理行动关闭 339 个非法网

球员中排在第 17 位。虽然乔普拉的莱德杯世界积分(Ryder

Figure 4: Correct recognition samples.

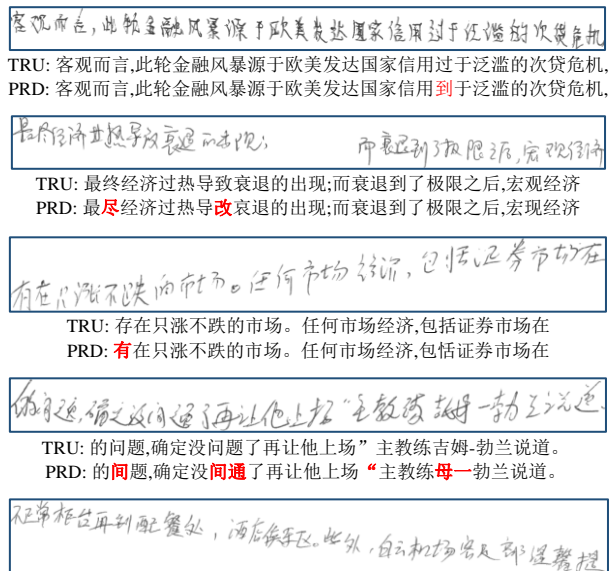

TRU: 客观而言,此轮金融风暴源于欧美发达国家信用过于泛滥的次贷危机,
PRD: 客观而言,此轮金融风暴源于欧美发达国家信用到于泛滥的次贷危机,

TRU: 最终经济过热导致衰退的出现;而衰退到了极限之后,宏观经济
PRD: 最尽经济过热导改衰退的出现;而衰退到了极限之后,宏现经济

TRU: 存在只涨不跌的市场。任何市场经济,包括证券市场在
PRD: 有在只涨不跌的市场。任何市场经济,包恬证券市场在

TRU: 的问题,确定没问题了再让他上场"主教练吉姆-勃兰说道。
PRD: 的间题,确定没间通了再让他上场"主教练母一勃兰说道。

TRU: 不正常柜台再到配餐处、酒店候车区。此外,白云机场客运部温馨提
PRD: 不常柜台再到配餐处,酒候车。些外,白云机客运部湿馨提

Figure 5: Misrecognized samples. TRU is the ground truth and PRD is the predicted result.